\documentclass[final]{cvpr}

\usepackage{times}
\usepackage{epsfig}
\usepackage{graphicx}
\usepackage{amsmath}
\usepackage{amssymb}
\usepackage{float}
\usepackage{booktabs}
\usepackage{multirow}
\usepackage{rotating}
\usepackage{bbding}
\usepackage{enumitem}
\usepackage{bm}
\usepackage{colortbl} % \cellcolor
\definecolor{mygray}{gray}{0.6}
\definecolor{mygray-bg}{gray}{0.9}
\usepackage[pagebackref=true,breaklinks=true,colorlinks,bookmarks=false]{hyperref}
\usepackage{soul}
\usepackage[font=small,labelfont=bf]{caption}
\usepackage{footmisc}

\newcommand\blfootnote[1]{%
  \begingroup
  \renewcommand\thefootnote{}\footnote{#1}%
  \addtocounter{footnote}{-1}%
  \endgroup
}

\usepackage{subfig}
\usepackage{array}

\newcolumntype{x}[1]{>{\centering\arraybackslash}p{#1pt}}
\newcommand{\tablestyle}[2]{\setlength{\tabcolsep}{#1}\renewcommand{\arraystretch}{#2}\centering\footnotesize}

\usepackage[dvipsnames]{xcolor}
\definecolor{mypink}{cmyk}{0, 0.7808, 0.4429, 0.1412}
\definecolor{mygreen}{rgb}{0.0, 0.7, 0.0}
\definecolor{myblue}{rgb}{0.0, 0.72, 0.92}

\def\cvprPaperID{5847} % *** Enter the CVPR Paper ID here
\def\confYear{CVPR 2022}

\begin{document}

%%%%%%%%% TITLE
\title{The Devil is in the Labels: \\
Noisy Label Correction for Robust Scene Graph Generation}

\author{
    Lin Li$^1$,
    Long Chen$^{2\dagger}$,
    Yifeng Huang$^1$,
    Zhimeng Zhang$^1$,
    Songyang Zhang$^3$,
    Jun Xiao$^1$
    \\
    $^1$Zhejiang University, \;\; $^2$Columbia University, \;\; $^3$University of Rochester \\
 {\tt\small\{mukti,yfhuang,zhimeng,junx\}@zju.edu.cn} ~
 {\tt\small zjuchenlong@gmail.com} ~
 {\tt\small szhang83@ur.rochester.edu}
}

\twocolumn[{%
    \maketitle
    \begin{figure}[H]
    \hsize=\textwidth
    \centering
    \includegraphics[width=17.5cm]{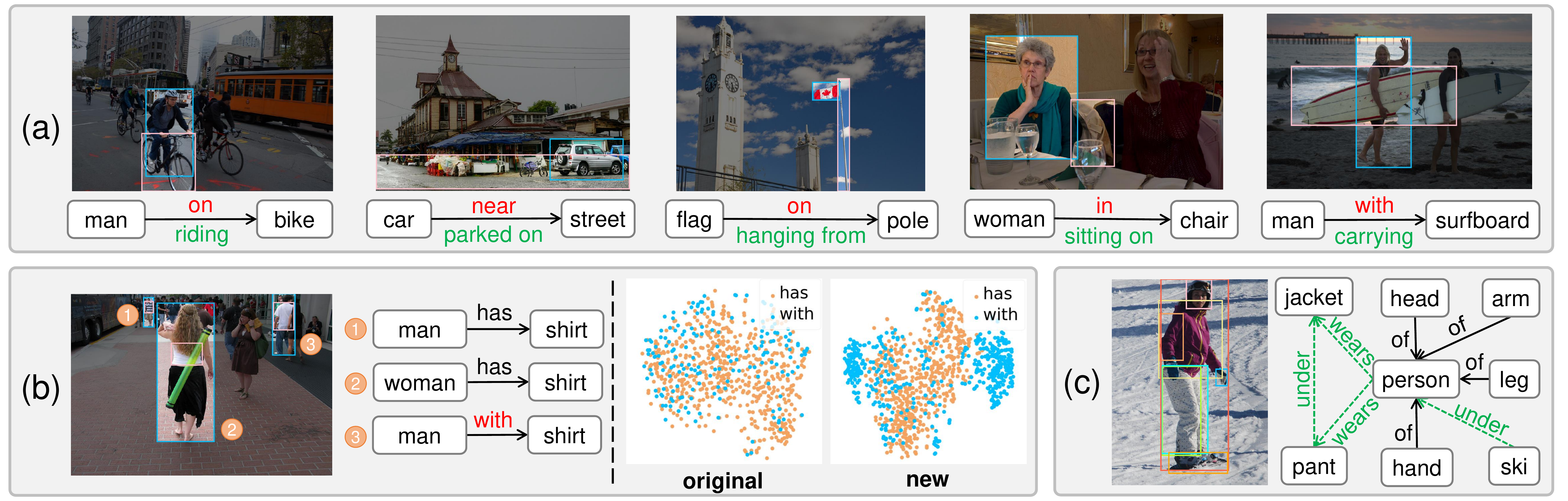}
    \vspace{-2em}
    \caption{Illustration of three types of noisy annotations in SGG datasets (take VG as an example). (a) \textbf{Common-prone}: For some triplets, the annotators tend to select less informative coarse-grained predicates (\textcolor{red}{\textbf{red}}) instead of fine-grained ones (\textcolor{mygreen}{\textbf{green}}). The subject and object for each triplet are denoted by \textcolor{myblue}{\textbf{blue}} and \textcolor{mypink}{\textbf{pink}} boxes, respectively. (b) \textbf{Synonym-random}: For some triplets, annotators usually randomly choose one predicate from the several synonyms (\eg, \texttt{has} and \texttt{with} are synonyms for $\langle$\texttt{man/woman}-\texttt{shirt}$\rangle$). \emph{Original:} The t-SNE visualization of original triplets $\langle$\texttt{man}-\texttt{has/with}-\texttt{shirt}$\rangle$ features. For brevity, we randomly sample part of triplets for each type. \emph{New}: The t-SNE visualization of same triplets after NICE. (c) \textbf{Negative}: Some negative triplets may not be \texttt{background} (\textcolor{mygreen}{\textbf{green}} dash arrows).
    }
    \label{fig:motivation}
\end{figure}
}]

\pagestyle{empty}  % no page number for the second and the later pages
\thispagestyle{empty} % no page number for the first page

%%%%%%%%% ABSTRACT
\begin{abstract}
\blfootnote{$^\dagger$ Corresponding author. This work started when Long Chen at ZJU.}
\blfootnote{Codes available at: \href{https://github.com/muktilin/NICE}{https://github.com/muktilin/NICE}.}
Unbiased SGG has achieved significant progress over recent years. However, almost all existing SGG models have overlooked the ground-truth annotation qualities of prevailing SGG datasets, \ie, they always assume: 1) all the manually annotated positive samples are equally correct; 2) all the un-annotated negative samples are absolutely background. In this paper, we argue that both assumptions are inapplicable to SGG: there are numerous ``noisy" ground-truth predicate labels that break these two assumptions, and these noisy samples actually harm the training of unbiased SGG models. To this end, we propose a novel model-agnostic \textbf{N}o\textbf{I}sy label \textbf{C}orr\textbf{E}ction strategy for SGG: \textbf{NICE}. NICE can not only detect noisy samples but also reassign more high-quality predicate labels to them. After the NICE training, we can obtain a cleaner version of SGG dataset for model training. Specifically, NICE consists of three components: negative Noisy Sample Detection (Neg-NSD), positive NSD (Pos-NSD), and Noisy Sample Correction (NSC). Firstly, in Neg-NSD, we formulate this task as an out-of-distribution detection problem, and assign pseudo labels to all detected noisy negative samples. Then, in Pos-NSD, we use a clustering-based algorithm to divide all positive samples into multiple sets, and treat the samples in the noisiest set as noisy positive samples. Lastly, in NSC, we use a simple but effective weighted KNN to reassign new predicate labels to noisy positive samples. Extensive results on different backbones and tasks have attested to the effectiveness and generalization abilities of each component of NICE.
\end{abstract}

%%%%%%%%% BODY TEXT
\section{Introduction}

Scene Graph Generation (SGG), \ie, detecting all object instances and their pairwise visual relations, is a crucial step towards comprehensive visual scene understanding. In general, each scene graph is a visually-grounded graph, where each node and edge refer to an object and visual relation, respectively. Recently, with the release of several large-scale SGG benchmarks (\eg, Visual Genome (VG)~\cite{krishna2017visual}) and advanced object detectors~\cite{ren2015faster,carion2020end,wang2022crossformer}, SGG has received unprecedented attention~\cite{gao2022classification}. However, due to the compositional nature of pairwise visual relations, the number distributions of different triplets in SGG datasets are much more imbalanced (\ie, long-tailed) than other recognition tasks. Accordingly, the performance of many state-of-the-art SGG models~\cite{zellers2018neural, chen2019counterfactual, tang2019learning, lin2020gps} degrades significantly on the \emph{tail} categories\footnote{For brevity, we directly use ``tail", ``body", and ``head" categories to represent the predicate categories in the tail, body, and head parts of the number distributions of different predicates in SGG datasets, respectively.} compared to the \emph{head} categories counterparts.

Currently, the mainstream solutions to mitigate the long-tailed problem in SGG can be coarsely categorized into two types: 1) \textit{Re-balancing strategy}: It utilizes class-aware sample re-sampling or loss re-weighting to balance the proportions of different predicate categories in the network training. The former attempts to balance the number of training samples in instance-level\footnote{We use ``instance" to denote an instance of visual relation triplet, and we also use ``sample" to represent the triplet instance interchangeably. \label{footnote:sample}} or image-level~\cite{li2021bipartite}, and the latter leverages prior commonsense knowledge (\eg, frequency of predicates~\cite{lin2017focal}, predicate correlations~\cite{yan2020pcpl}, or rule-based predicate priority~\cite{lin2020gps, knyazev2020graph}) to re-weight the contributions of different categories in loss calculations. 2) \textit{Biased-model-based strategy}: It inferences debiased predictions from pretrained biased SGG models. For instance, using counterfactual causality to disentangle frequency biases~\cite{tang2020unbiased}, deriving more balanced loss weights for different predicates~\cite{yu2021cogtree}, or adjusting the probabilities of predicate predictions~\cite{chiou2021recovering}.

Although these methods have dominated performance on debiasing metrics (\eg, mean Recall@K), it is worth noting that almost all existing models have taken two plausible assumptions about the ground-truth annotations for granted:
\begin{itemize}[leftmargin=0.4cm]
    \item[] \textbf{Assumption 1:} \emph{All the manually annotated positive samples are equally correct.}

    \item[] \textbf{Assumption 2:} \emph{All the un-annotated negative samples are absolutely background.}
\end{itemize}

For the first assumption, by ``equally", we mean that the confidence (or quality) of annotated ground-truth predicate label for each positive sample\footref{footnote:sample} is the same as others, \ie, all positive predicate labels are of high quality. Unfortunately, unlike other close-set classification tasks where each sample has only a unique ground-truth label, a subject-object pair in SGG sometimes has multiple reasonable predicates. This phenomenon has led to two inevitable annotation characteristics in SGG datasets: 1) \textbf{\emph{Common-prone}}: When these reasonable relations are in different semantic granularities, the annotators tend to select the most common predicate (or coarse-grained) as ground-truth. As shown in Figure~\ref{fig:motivation}(a), both \texttt{riding} and \texttt{on} are ``reasonable" for \texttt{man} and \texttt{bike}, but the annotated ground-truth predicate is less informative \texttt{on} instead of more convincing \texttt{riding}. And this characteristic is very common in SGG datasets (more examples in Figure~\ref{fig:motivation}(a)). 2) \textbf{\emph{Synonym-random}}: When these reasonable relations are synonymous for the subject-object pair, the annotators usually randomly choose one predicate as ground-truth, \ie, the annotations for some similar visual patterns are inconsistent. For example, in Figure~\ref{fig:motivation}(b), both \texttt{has} and \texttt{with} denote ``be dressed in" for \texttt{man}/\texttt{woman} and \texttt{shirt}, but the ground-truth annotations are inconsistent even in the same image. We further visualize thousands of sampled instances of $\langle$\texttt{man}-\texttt{has} / \texttt{with}-\texttt{shirt}$\rangle$ in VG, and these instances are all randomly distributed in the feature space (cf. Figure~\ref{fig:motivation}(b)). Thus, \emph{we argue that all the positive samples are NOT equally correct, i.e., a part of positive samples are not high-quality --- their labels can be more fine-grained (cf. common-prone) or more consistent (cf. synonym-random).}

For the second assumption, although all SGG works have agreed that visual relations in existing datasets are always sparsely identified and annotated~\cite{lu2016visual} (Figure~\ref{fig:motivation}(c)), almost all of them still train their models by regarding all the un-annotated pairs as \texttt{background}, \ie, there is no visual relation between the subject and object. In contrast, \emph{we argue that all negative samples are NOT absolutely background, i.e., a part of negative samples are not high-quality --- they are actually foreground with missing annotations.}

In this paper, we try to get rid of these two questionable assumptions, and reformulate SGG as a noisy label learning problem. To the best of our knowledge, we are the first work to take a deep dive into the ground-truth annotation qualities of both positive and negative samples in SGG. Specifically, we propose a novel model-agnostic \textbf{N}o\textbf{I}sy label \textbf{C}orr\textbf{E}ction strategy, dubbed as \textbf{NICE}. NICE can not only detect numerous \emph{noisy} samples, but also reassign more high-quality predicate labels to them. By ``noisy", we mean that these samples break these two assumptions. After the NICE training, we can obtain a cleaner version of dataset for SGG training. Particularly, we can: 1) increase the number of fine-grained predicates (\emph{common-prone}); 2) decrease annotation inconsistency among similar visual patterns (\emph{synonym-random}); 3) increase the number of positive samples (\emph{assumption 2}).

NICE consists of three components: negative noisy sample detection (Neg-NSD), positive noisy sample detection (Pos-NSD), and noisy sample correction (NSC). Firstly, in Neg-NSD, we reformulate the negative NSD as an out-of-distribution (OOD) detection problem, \ie, regarding all the positive samples as in-distribution (ID) training data, and all the un-annotated negative samples as OOD test data. In this way, we can detect the missing annotated (ID) samples with pseudo labels. Then, in Pos-NSD, we use a clustering-based algorithm to divide all positive samples (including the outputs of Neg-NSD) into multiple sets, and regard samples in the noisiest set as noisy positive samples. The clustering results are based on the local density of each sample. Lastly, in NSC, we use a simple but effective weighted KNN to reassign new predicate labels to all noisy positive samples.

We evaluate NICE on the most prevalent SGG benchmark: VG~\cite{krishna2017visual}. Since NICE only focuses on refining noisy annotations of the dataset, it can be seamlessly incorporated into any SGG architecture to boost their performance. Extensive ablations have attested to the effectiveness and generalization abilities of each component of NICE.

In summary, we make three contributions in this paper:
\vspace{-0.5em}
\begin{enumerate}[leftmargin=4mm]
    \itemsep-0.4em

    \item We are the first to reformulate SGG as a noisy label learning problem, and point out the two plausible assumptions are not applicable for SGG, \ie, the devil is in the labels.

    \item We propose a novel model-agnostic strategy NICE. Extensive ablations on several baselines, tasks, and metrics have demonstrated its excellent generalization abilities.

    \item Each part of NICE can serve as an independent plug-and-play module to improve SGG annotation qualities\footnote{For example, the Pos-NSD can help models gain good results with much fewer training samples, and the Neg-NSD can generate plentiful unseen reasonable visual triplets. More details are left in Sec.~\ref{sec:4} and appendix.}.
\end{enumerate}

%-------------------------------------------------------------------
\begin{figure*}[!t]
  \centering
  \includegraphics[width=\linewidth]{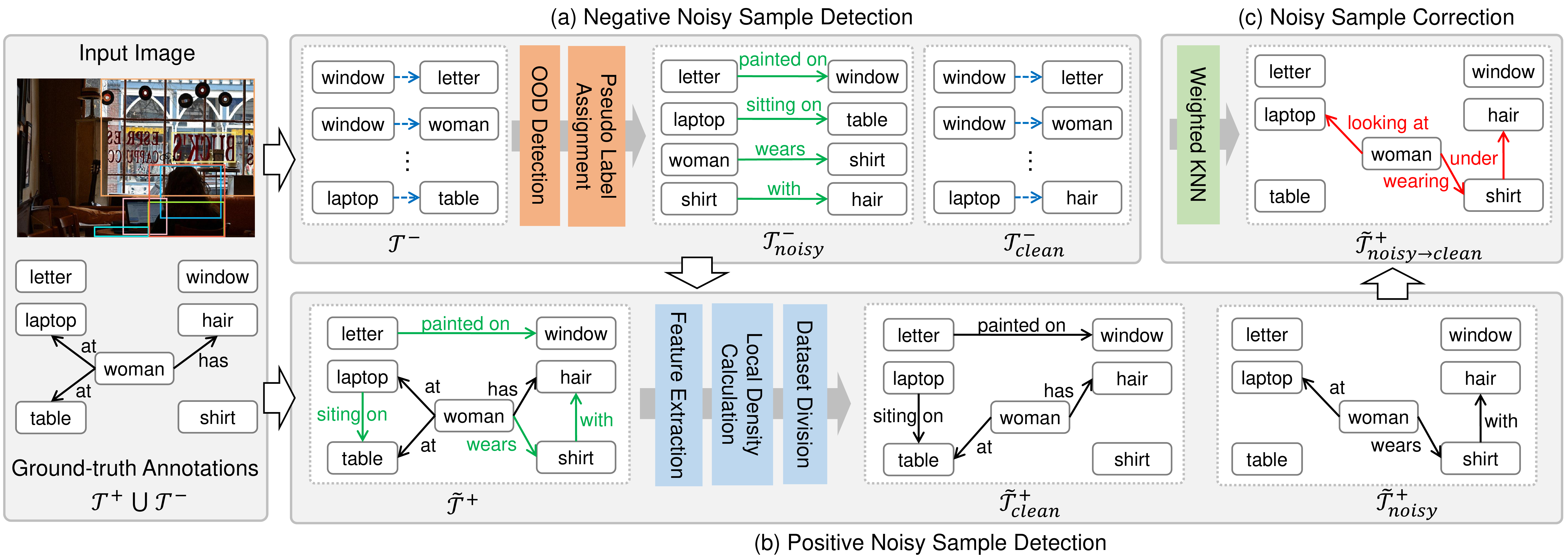}
  \vspace{-2em}
  \caption{The pipeline of NICE (take an image from VG as an example). (a) \textbf{Neg-NSD}: Given all negative triplets (\textcolor{cyan}{\textbf{blue}} dash arrows), the OOD detection model detects missing annotated triplets ($\mathcal{T}^-_{\text{noisy}}$) and assigns pseudo labels to them (\textcolor{mygreen}{\textbf{green}} predicates). (b) \textbf{Pos-NSD}: Given the new composed positive triplet set ($\widetilde {\mathcal{T}}^+$), Pos-NSD detects all noisy positive samples ($\widetilde {\mathcal{T}}^+_{\text{noisy}}$). (c) \textbf{NSC}: NSC reassigns more high-quality predicate labels to all noisy positive samples (\textcolor{red}{\textbf{red}} predicates). Finally, we obtain a new cleaner version of ground-truth annotations.}
  \label{fig:framework}
\end{figure*}
%-------------------------------------------------------------------

\section{Related Work} \label{related_work}

\noindent\textbf{Scene Graph Generation.} SGG aims to transform visual data into semantic graph structures. Early methods~\cite{lu2016visual, zhang2017visual} always ignore the visual context, \ie, they regard each object as an individual and predict pairwise relations directly. Subsequent SGG works start to utilize the overlooked visual context by resorting to different advanced techniques, \eg, message passing~\cite{xu2017scene, li2017vip, zellers2018neural, chen2019knowledge, chen2019counterfactual}, or tree/graph structure modelling~\cite{yang2018graph, tang2019learning}. Recently, unbiased SGG has drawn unprecedented attention, \ie, they focus on the performance gaps in different predicate categories. As above mentioned, existing unbiased SGG models can be categorized into: re-balancing strategy~\cite{li2021bipartite,lin2017focal,yan2020pcpl,lin2020gps, knyazev2020graph} and biased-model-based strategy~\cite{tang2020unbiased,yu2021cogtree, chiou2021recovering,guo2021general}. Different from existing SGG works, we are the first to explicitly refine the original noisy ground-truth annotations on SGG datasets. Although some previous works also have discussed the issue of sparse annotations~\cite{wang2020tackling, chiou2021recovering} or semantic imbalance~\cite{guo2021general}, they still heavily rely on these original noisy annotations in model training.

\noindent\textbf{Learning with Noisy Labels.}
Existing noisy label learning methods can be roughly divided into two categories: 1) Utilizing an explicit or implicit noise model to estimate the distribution of noisy and clean labels, and then deleting or correcting the noise samples. These models can be: neural networks~\cite{goldberger2016training, jiang2018mentornet, lee2018cleannet, ren2018learning}, conditional random field~\cite{vahdat2017toward} or knowledge graphs~\cite{li2017learning}. However, they always need abundant clean samples for training, which is always inapplicable for many noisy label learning datasets. 2) Constructing a more balanced loss function to reduce the influence of noisy samples~\cite{ma2018dimensionality, zhang2018generalized, wang2019symmetric, xu2019l_dmi}. In this paper, we are the first to formulate SGG as a noisy label learning problem, and propose a novel noisy sample detection and correction strategy.

\section{Approach} \label{approach}

Given an image dataset $\bm{\mathcal{I}}$, SGG task aims to convert each image $\mathcal{I}_i \in \bm{\mathcal{I}}$ into a graph $\mathcal{G}_i = \{ \mathcal{N}_i,  \mathcal{E}_i \}$, where $\mathcal{N}_i$ and $\mathcal{E}_i$ denote the node set (\ie, objects) and edge set (\ie, visual relations) of image $\mathcal{I}_i$, respectively. In general, each graph $\mathcal{G}_i$ can also be viewed as a set of visual relation triplets (\ie, $\langle$\texttt{subject}-\texttt{predicate}-\texttt{object}$\rangle$), denoted as $\mathcal{T}_i$. For each triplet set $\mathcal{T}_i$, we can further divide it into two subsets: $\mathcal{T}^+_i$ and $\mathcal{T}^-_i$, where $\mathcal{T}^+_i$ denotes all the annotated positive triplets (or samples) in image $\mathcal{I}_i$, and $\mathcal{T}^-_i$ denotes all the un-annotated negative triplets in image $\mathcal{I}_i$. Analogously, we use $\bm{\mathcal{T}}^+ = \{\mathcal{T}^+_i\}$ and $\bm{\mathcal{T}}^- = \{\mathcal{T}^-_i\}$ to represent all positive and negative triplets in the whole dataset $\bm{\mathcal{I}}$.

The whole pipeline of NICE is illustrated in Figure~\ref{fig:framework}\footnote{In Figure~\ref{fig:framework}, we use a single image as input for a clear illustration. In real experiments, we directly process the whole dataset in each module. \label{footnote:fig2}}. In this section, we sequentially introduce each part of NICE, including negative noisy sample detection (Neg-NSD), positive NSD (Pos-NSD), and noisy sample correction (NSC). Specifically, given an image and its corresponding ground-truth triplet annotations (\ie, $\mathcal{T}^+ \bigcup \mathcal{T}^-$)\footnote{For brevity, we omit the subscripts $i$ for image $\mathcal{I}_i$.}, we first use the Neg-NSD to detect all possible noisy negative samples, \ie, missing annotated foreground triplets. The $\mathcal{T}^-$ can be divided into  $\mathcal{T}^-_{\text{clean}}$ and $\mathcal{T}^-_{\text{noisy}}$. Meanwhile, Neg-NSD will assign pseudo positive predicate labels for all samples in $\mathcal{T}^-_{\text{noisy}}$ (\eg, \texttt{painted on} for $\langle$\texttt{letter}-\texttt{window}$\rangle$ in Figure~\ref{fig:framework}). The $\mathcal{T}^-_{\text{noisy}}$ with pseudo positive labels and original $\mathcal{T}^+$ compose a new positive set $\widetilde {\mathcal{T}}^+$. Then, we use the Pos-NSD to detect all possible noisy positive samples in $\widetilde {\mathcal{T}}^+$, \ie, the positive samples which suffer from either common-prone or synonym-random characteristics (\eg, \texttt{at} for $\langle$\texttt{women}-\texttt{laptop}$\rangle$ in Figure~\ref{fig:framework}). Similarly, $\widetilde{\mathcal{T}}^+$ can be divided into  $\widetilde{\mathcal{T}}^+_{\text{clean}}$ and $\widetilde{\mathcal{T}}^+_{\text{noisy}}$. Next, we use NSC to reassign more high-quality predicate labels to all samples in $\widetilde{\mathcal{T}}^+_{\text{noisy}}$, denote as $\widetilde{\mathcal{T}}^+_{\text{noisy} \to \text{clean}}$. Lastly, after processing the ground-truth triplet annotations of all images, we can obtain a cleaner version of dataset ($\widetilde{\bm{\mathcal{T}}}^+_{\text{clean}} \bigcup \widetilde{\bm{\mathcal{T}}}^+_{\text{noisy} \to \text{clean}} \bigcup \bm{\mathcal{T}}^-_{\text{clean}}$) for SGG training.

\subsection{Negative Noisy Sample Detection (Neg-NSD)}
In this module, we aim to discover all possible \emph{noisy} negative samples, \ie, missing annotated visual relation triplets. Meanwhile, Neg-NSD also assigns a pseudo positive predicate label for each detected noisy negative sample. Due to the nature of missing annotations in existing negative samples, it is difficult to directly train and evaluate a binary classifier based on these noisy samples. To this end, we propose to formulate the negative noisy sample detection as an out-of-distribution (OOD) detection problem~\cite{hendrycks2016baseline}. Specifically, we regard all annotated positive samples as in-distribution (ID) training data, and all un-annotated negative samples as OOD test data. The Neg-NSD is built on top of a plain SGG model (denoted as $\mathtt{F}_{\text{sgg}}^n$), but it is trained with only the annotated positive samples $\bm{\mathcal{T}}^+$. In the inference stage, Neg-NSD will predict a score of being foreground and a pseudo positive predicate category for each triplet $t^-_i \in \bm{\mathcal{T}}^-$.

Following existing OOD detection methods~\cite{devries2018learning}, we also utilize a confidence-based model, \ie, Neg-NSD consists of two network output branches: 1) a classification branch to predict a probability distribution $\bm{p}$ over all positive predicate categories, and 2) a confidence branch to predict a confidence score $c \in [0, 1]$, which indicates the confidence of being an ID category (foreground). In the inference stage, for each sample $t^-_i$, if its confidence score $c_i$ is larger than a threshold $\theta$, we then regard this negative sample as a noisy negative sample, \ie, the detection function $g(\cdot)$ is as:
\begin{equation} \label{eq:1}
g(t^-_i) = \left\{
\begin{array}{l}
1, \quad c_i \ge \theta \\
0, \quad c_i < \theta.
\end{array} \right.
\end{equation}
When $g(t^-_i) = 1$, the pseudo label of $t^-_i$ is directly derived from the classification branch, \ie, $\arg \max (\bm{p}_i)$. Since the predicted average confidence scores vary considerably for different predicate categories, we set different thresholds for head, body, and tail categories. (More details are in Sec.~\ref{sec:4}.)

\textbf{Training of Neg-NSD.} To train the classification branch and confidence branch, we combine predicted probabilities $\bm{p}_i$ and corresponding target probability distribution $\bm{y}_i$, \ie,
\begin{equation} \label{eq:2}
\bm{p'}_i = c_i \cdot {\bm{p}_i} + (1 - c_i) \cdot \bm{y}_i,
\end{equation}
where $\bm{p'}_i$ is the adjusted probabilities by confidence $c_i$. The motivation of Eq.~\eqref{eq:2} is that if the model is given a chance to ask for a hint of the ground-truth probability with some penalty, the model will definitely choose to ask for the hint if it is not confident about its output (\ie, $c_i$ is small). And the training objective for Neg-NSD consists of a weighted cross-entropy loss and a regularization penalty loss:
\begin{equation}
    \mathcal{L} = - \textstyle{\sum}_{j = 1} {{w_{j}}\log ({p'}_{ij}){{y}_{ij}} - \lambda \log (c_{i})},
\end{equation}
where $p'_{ij}$ and $y_{ij}$ are the $j$-th element of $\bm{p'}_i$ and $\bm{y}_i$, respectively. $w_j$ is the reciprocal of the frequency of the $j$-th predicate category, which mitigates the impact of the long-tail issues on confidence. The penalty loss is utilized to prevent the network from always choosing $c$ = 0 and using ground-truth probability distribution to minimize task loss.

\begin{figure}[!t]
  \centering
  \includegraphics[width=\linewidth]{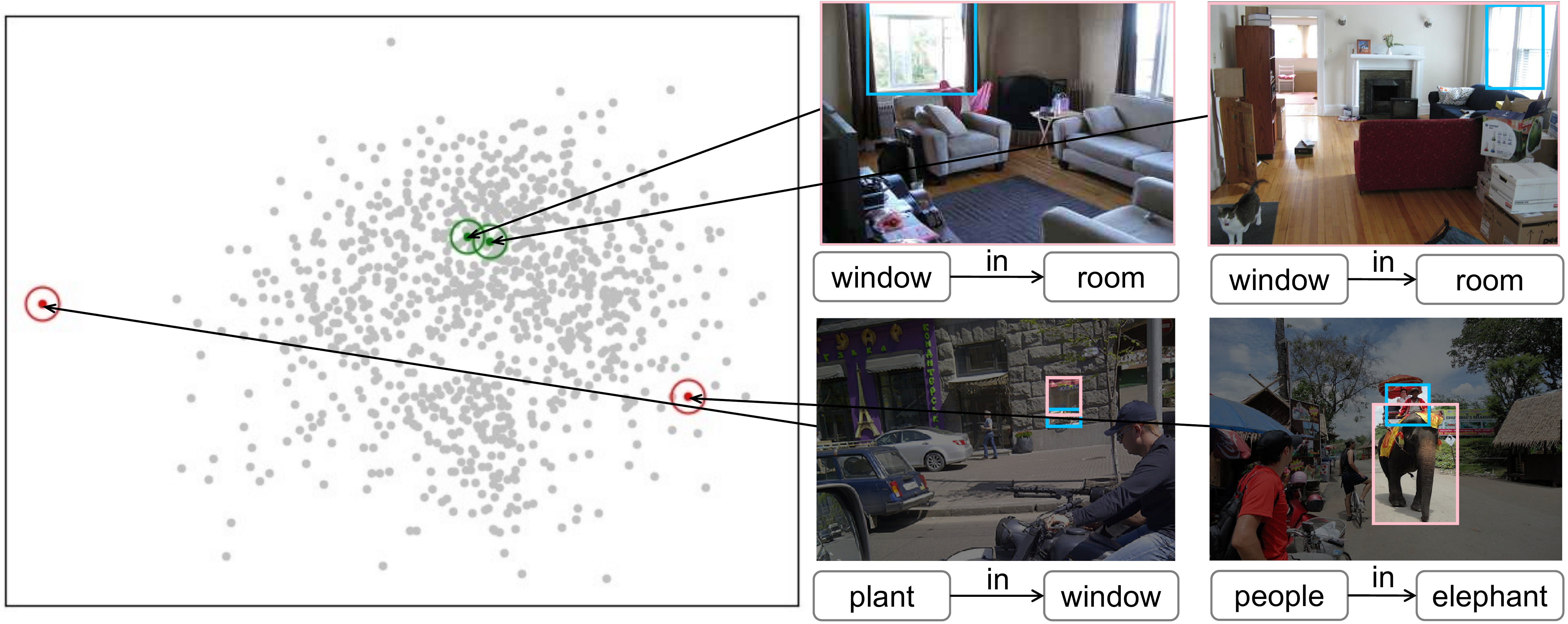}
  \vspace{-2em}
  \caption{\textbf{Left}: Multidimensional scaling visualization of features of randomly sampled triplets with predicate \texttt{in}. \textbf{Right}: Detected \textcolor{mygreen}{\textbf{clean}}  samples and \textcolor{red}{\textbf{noisy}} samples by Pos-NSD.}
  \label{fig:example}
\end{figure}

\subsection{Positive Noisy Sample Detection (Pos-NSD)}
\label{sec:3.2}
As shown in Figure~\ref{fig:framework}\footref{footnote:fig2}, the original positive set $\bm{\mathcal{T}}^+$ and outputs of the Neg-NSD (\ie, $\bm{\mathcal{T}}^-_{\text{noisy}}$) compose a new positive sample set ${\widetilde {\bm{\mathcal{T}}}^+}$. The Pos-NSD module aims to detect all \emph{noisy} samples in ${\widetilde{\bm{\mathcal{T}}}^+}$. In general, we use a clustering-based solution to divide all these positive samples into multiple subsets with different degrees of noise, and treat all samples in the noisiest subset as noisy positive samples. Intuitively, if a predicate label is consistent with other visually-similar samples of the same predicate category (\ie, visual features of these samples are close to each other), this predicate is more likely to be a clean sample, because these annotations are consistent with each others. Otherwise, it is likely to be a noisy sample. As shown in Figure~\ref{fig:example}, the two clean triplets $\langle$\texttt{window}-\texttt{in}-\texttt{room}$\rangle$ have more visually-similar neighbors than the noisy triplets (\eg, $\langle$\texttt{plant}-\texttt{in}-\texttt{window}$\rangle$).

\begin{figure}[!t]
  \centering
  \includegraphics[width=\linewidth]{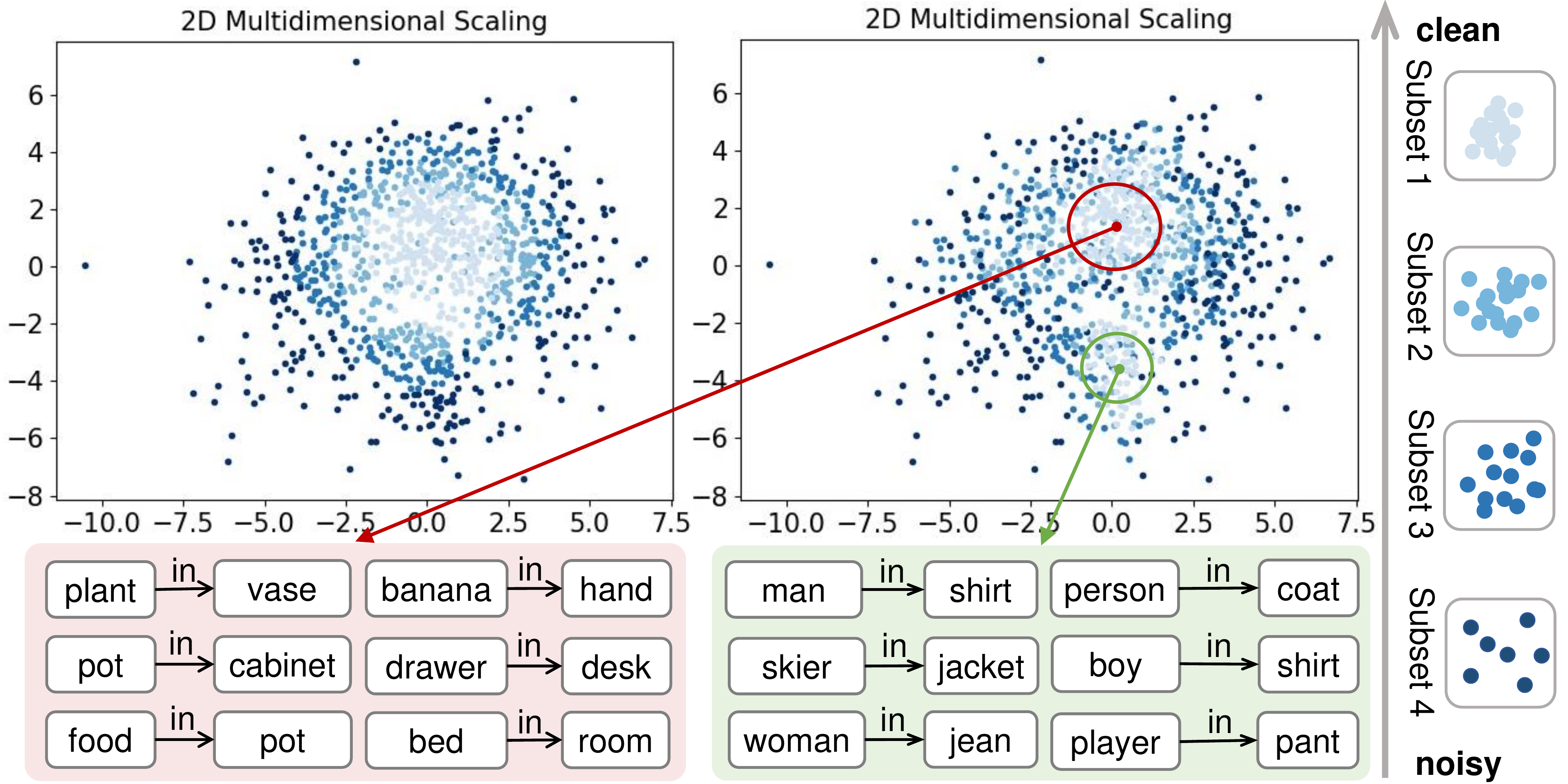}
  \vspace{-1.5em}
  \caption{\textbf{Above}: The multidimensional scaling visualization of features of randomly sampled triplets with predicate \texttt{in} with cutoff distance ranked at 50\% (left) and 1\% (right). \textbf{Below}: The triplet categories of the randomly sampled visual relation triplets from the corresponding \textcolor{red}{\textbf{red}} circle and \textcolor{mygreen}{\textbf{green}} circle.}
  \label{fig:subset}
\end{figure}

Based on these observations, we propose a local density based solution for positive noisy sample detection. Specifically, we utilize an off-the-shelf pretrained SGG model (denoted as $\mathtt{F}_{\text{sgg}}^p$) to extract all visual relation triplet features, and use $\bm{h}^k_i$ to represent the visual feature of $i$-th sample of predicate category $k$ (this sample is denoted as $t^k_i$). Then, we utilize a distance matrix $\bm{D}^k = (d^k_{ij})_{N \times N} \in \mathbb{R}^{N \times N}$ to measure the similarity between all positive samples of the same predicate $k$, and $d^k_{ij}$ is calculated by:
\begin{equation} \label{eq:4}
    d^k_{ij} = \left\| \bm{h}^k_i - \bm{h}^k_j \right\|^2,
\end{equation}
where $\|\cdot\|$ is the Euclidean distance. Thus, a smaller distance $d^k_{ij}$ means a relatively higher similarity between sample $t^k_i$ and sample $t^k_j$. Then following~\cite{rodriguez2014clustering}, we define the local density $\rho^k_i$ of each sample $t^k_i$ as the number of samples (within the same predicate category) whose similarity distance to sample $t^k_i$ are closer than a threshold $d^k_c$, \ie,
\begin{equation} \label{eq:5}
{\rho^k _i} = \textstyle\sum_j {\mathbf{1} (({d^k_c} - {d^k_{ij}})>0)},
\end{equation}
where $\mathbf{1}(\cdot)$ is the indicator function
and $d^k_c$ is the cutoff distance for predicate $k$, which is ranked at $\alpha\%$ of sorted $N \times N$ distances in $\bm{D}^K$ from small to large. Thus, a sample with higher local density $\rho$ means that this sample is more similar to the samples of the same predicate category. Analogously, samples with low local density $\rho$ are considered as noisy samples. Finally, we use an unsupervised K-means algorithm~\cite{hechenbichler2004weighted} to divide all data samples into multiple subsets with respect to different $\rho$ values, \ie, different degrees of noise~\cite{guo2018curriculumnet}. And all samples in the subset with the lowest $\rho$ are regarded as noisy positive samples (\ie, $\widetilde{\bm{\mathcal{T}}}^+_{\text{noisy}}$), and fed into the following NSC module for label correction.

\textbf{Influence of the Cutoff Distance $d^k_c$.} From Eq.~\eqref{eq:5}, we can observe that the distribution of local density $\rho$ is directly decided by the selection of the cutoff distance $d_c$ (or the hyperparameter $\alpha\%$). As shown in Figure~\ref{fig:subset}, when the cutoff distance ranked at 50\% and 1\%, local densities of samples diffuse outwards from large to small with one and two centers, respectively, \ie, a smaller cutoff distance (\eg, 1\% for $\alpha\%$) may divide the whole feature space into more clusters. Meanwhile, different predicate categories may contain variable types of semantic meanings. For example, in Figure~\ref{fig:subset}, predicate \texttt{in} of samples inside the red circle represents ``inside" (\eg, $\langle$\texttt{plant}-\texttt{in}-\texttt{vase}$\rangle$), while predicate \texttt{in} of the samples inside the green circle represents ``wearing" (\eg, $\langle$\texttt{man}-\texttt{in}-\texttt{shirt}$\rangle$). Thus, we set different cutoff distances to different categories. More details are in Sec.~\ref{sec:4}.

In addition, more detailed discussions about the influence of $d^k_c$ on the clustering results are left in the appendix.

\begin{figure}[!t]
  \centering
  \includegraphics[width=\linewidth]{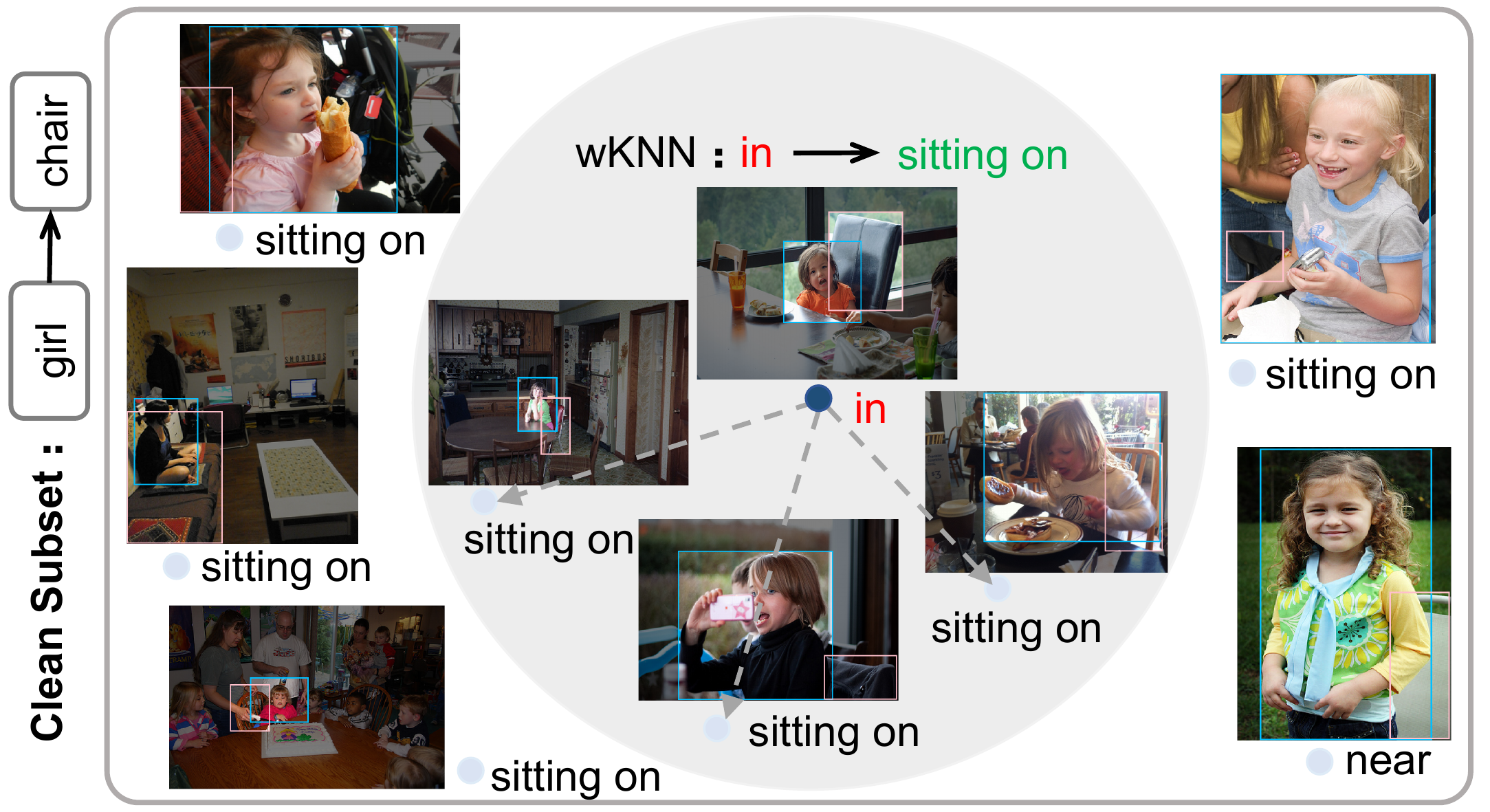}
  \vspace{-1.5em}
  \caption{The illustration of NSC. Dashed lines indicate the distances between the noisy sample and other samples in clean subset with \texttt{girl}-\texttt{chair}. wKNN replaces noisy \texttt{in} to \texttt{sitting on}.}
  \label{fig:correction}
\end{figure}

\subsection{Noisy Sample Correction (NSC)}

Given all the detected noisy positive samples from Pos-NSD, the NSC module aims to correct these noisy positive predicate labels. The motivation of our NSC is that the predicate label of a sample should be consistent with other visually-similar samples, especially for those samples with the same \texttt{subject} and \texttt{object} categories. For example, in Figure~\ref{fig:correction}, for the noisy sample $\langle$\texttt{girl}-\texttt{in}-\texttt{chair}$\rangle$, we can retrieval all other samples with the same $\langle$\texttt{girl}-\texttt{chair}$\rangle$, and find most of the visually-similar samples are annotated as $\langle$\texttt{girl}-\texttt{sitting on}-\texttt{chair}$\rangle$. Thus, we use the simple but effective weighted K-Nearest Neighbor (wKNN) algorithm to derive the most possible labels for noisy positive samples. The wKNN assigns larger weights to the closest samples and smaller weights to the ones that are far away. Specifically, let $N(i)$ be the set of $K$ neighbors of sample $t_i$, then the new assigned label for $t_i$ is:
\begin{equation}
    {r_i^{'}} = \arg {\max _v}\sum\nolimits_{t_j \in N(i)} {{w_{ij}} \cdot \mathbf{1} (v = {r_j})},
\end{equation}
where $v$ is a predicate category, $r_j$ is predicate label of $t_j$, and $\mathbf{1}(\cdot)$ is an indicator function. The weight $w_{ij}$ is assigned to each neighbor, defined as $a \cdot \exp(- \frac{(d_{ij} - b)^2}{2c^2})$. The $d_{ij}$ is the Euclidean distance between $\bm{h}_i$ and $\bm{h}_j$ (cf. Eq.~\eqref{eq:4}), and $a$, $b$, $c$ are hyperparameters. It is worth noting that since we only consider the samples with the same subject-object categories, we can solve NSC with fast inference speed. If the new label is the same as the old, no label will be assigned.

\addtolength{\tabcolsep}{-2pt}
\begin{table*}[tbp]
    \centering
        \scalebox{0.95}{
            \begin{tabular}{c| l |  c c  |c | c  c  |c |  c c  |c}
                \hline
                \multirow{2}{*}{B} &
                \multirow{2}{*}{Models} & \multicolumn{3}{c|}{PredCls} & \multicolumn{3}{c|}{SGCls} & \multicolumn{3}{c}{SGGen} \\
                \cline{3-11}
                & & \small{mR@50/100} & \small{R@50/100} & \small{Mean}
               & \small{mR@50/100} & \small{R@50/100} & \small{Mean}
               & \small{mR@50/100} & \small{R@50/100} & \small{Mean} \\
                \hline
               \multirow{4}{*}{\begin{sideways}VGG-16\end{sideways}}
                &
                Motif~\cite{zellers2018neural} $_{\textit{CVPR'18}}$
                & 14.0 / 15.3 & 65.2 / 67.1	& 40.4
                & 7.7  / 8.2  & 35.8 / 36.5 & 22.1
                & 5.7  / 6.6  & 27.2 / 30.3 & 17.5 \\
                & VCTree~\cite{tang2019learning}$_{\textit{CVPR'19}}$
                & 17.9 / 19.4 & 66.4 / 68.1	& 43.0
                & 10.1 / 10.8 & 38.1 / 38.8 & 24.5
                & 6.9  / 8.0  & 27.9 / 31.3 & 18.5 \\
                & KERN~\cite{chen2019knowledge}$_{\textit{CVPR'19}}$
                & 17.7 / 19.2 & 65.8 / 67.6	& 42.6
                & 9.4  / 10.0 & 36.7 / 37.4 & 23.4
                & 6.4  / 7.3  & 29.8 / 27.1 & 17.7 \\
                & PCPL~\cite{yan2020pcpl}$_{\textit{MM'20}}$
                & 35.2 / 37.8 & 50.8 / 52.6	& 44.1
                & 18.6 / 19.6 & 27.6 / 28.4 & 23.6
                & 9.5  / 11.7 & 14.6 / 18.6 & 13.6 \\
                \hline
                \multirow{18}{*}{\begin{sideways}X-101-FPN\end{sideways}}
                & MSDN~\cite{li2017scene}$_{\textit{ICCV'17}}$
                & 15.9 / 17.5 & 64.6 / 66.6 & 41.2
                & 9.3  / 9.7  & 38.4 / 39.8 & 24.3
                & 6.1  / 7.2  & 31.9 / 36.6 & 20.5 \\
                & G-RCNN~\cite{yang2018graph}$_{\textit{ECCV'18}}$
                & 16.4 / 17.2 & 64.8 / 66.7 & 41.3
                & 9.0  / 9.5  & 38.5 / 37.0 & 23.5
                & 5.8  / 6.6  & 29.7 / 32.8 & 18.7 \\
                & BGNN~\cite{li2021bipartite}$_{\textit{CVPR'21}}$
                & 30.4 / 32.9 & 59.2 / 61.3 & 45.9
                & 14.3 / 16.5 & 37.4 / 38.5 & 26.7
                & 10.7 / 12.6 & 31.0 / 35.8 & 22.5 \\
                & DT2-ACBS~\cite{desai2021learning}$_{\textit{ICCV'21}}$
                & 35.9 / 39.7 & 23.3 / 25.6 & 31.1
                & 24.8 / 27.5 & 16.2 / 17.6 & 21.5
                & 22.0 / 24.4 & 15.0 / 16.3 & 19.4 \\
                \cline{2-11}
                & Motifs~~\cite{zellers2018neural}$_{\textit{CVPR'18}}$
                & 16.5 / 17.8 & \textcolor{red}{\textbf{65.5}} / \textcolor{red}{\textbf{67.2}} & \textcolor{blue}{\textbf{41.8}}
                & 8.7  / 9.3  & \textcolor{red}{\textbf{39.0}} / \textcolor{red}{\textbf{39.7}} & \textcolor{blue}{\textbf{24.2}}
                & 5.5  / 6.8  & \textcolor{red}{\textbf{32.1}} / \textcolor{red}{\textbf{36.9}} & \textcolor{blue}{\textbf{20.3}} \\
               &  ~~+TDE~\cite{tang2020unbiased}$_{\textit{CVPR'20}}$
                & 24.2 / 27.9 & 45.0 / 50.6 & 36.9
                & 13.1 / 14.9 & 27.1 / 29.5 & 21.2
                & 9.2  / 11.1 & 17.3 / 20.8 & 14.6 \\
                & ~~+PCPL~\cite{yan2020pcpl}$_{\textit{MM'20}}$
                & 24.3 / 26.1 & 54.7 / 56.5 & 40.4
                & 12.0 / 12.7 & \textcolor{blue}{\textbf{35.3}} / \textcolor{blue}{\textbf{36.1}} & 24.0
                & 10.7 / 12.6 & 27.8 / 31.7 & 20.7 \\
                & ~~+CogTree~\cite{yu2021cogtree}$_{\textit{IJCAI'21}}$
                & 26.4 / 29.0 & 35.6 / 36.8	& 32.0
                & 14.9 / 16.1 & 21.6 / 22.2 & 18.7 	
                & 10.4 / 11.8 & 20.0 / 22.1 & 16.1 \\
                & ~~+DLFE~\cite{chiou2021recovering}$_{\textit{MM'21}}$
                & 26.9 / 28.8 & 52.5 / 54.2 & 40.6
                & 15.2 / 15.9 & 32.3 / 33.1 & 24.1
                & 11.7 / 13.8 & 25.4 / 29.4 & 20.1 \\
                & ~~+BPL-SA~\cite{guo2021general}$_{\textit{ICCV'21}}$
                & \textcolor{blue}{\textbf{29.7}} / \textcolor{blue}{\textbf{31.7}} & 50.7 / 52.5 & 41.2
                & \textcolor{blue}{\textbf{16.5}} / \textcolor{blue}{\textbf{17.5}} & 30.1 / 31.0 & 23.8
                & \textcolor{red}{\textbf{13.5}} / \textcolor{red}{\textbf{15.6}} & 23.0 / 26.9 & 19.8 \\
                & \cellcolor{mygray-bg}{~~+\textbf{NICE (ours)}}
                &  \cellcolor{mygray-bg}{\textcolor{red}{\textbf{29.9}}} / \cellcolor{mygray-bg}{\textcolor{red}{\textbf{32.3}}} &  \cellcolor{mygray-bg}\textcolor{blue}{\textbf{55.1}} / \cellcolor{mygray-bg}\textcolor{blue}{\textbf{57.2}} &  \cellcolor{mygray-bg}{\textcolor{red}{\textbf{43.6}}}
                &  \cellcolor{mygray-bg}{\textcolor{red}{\textbf{16.6}}} /  \cellcolor{mygray-bg}{\textcolor{red}{\textbf{17.9}}} &  \cellcolor{mygray-bg}{33.1} / \cellcolor{mygray-bg}{34.0} &  \cellcolor{mygray-bg}{\textcolor{red}{\textbf{25.4}}}
                &  \cellcolor{mygray-bg}\textcolor{blue}{\textbf{12.2}} /  \cellcolor{mygray-bg}\textcolor{blue}{\textbf{14.4}} &  \cellcolor{mygray-bg}\textcolor{blue}{\textbf{27.8}} / \cellcolor{mygray-bg}\textcolor{blue}{\textbf{31.8}} &  \cellcolor{mygray-bg}{\textcolor{red}{\textbf{21.6}}} \\

                \cline{2-11}
                & VCTree~\cite{tang2019learning}$_{\textit{CVPR'19}}$
                & 17.1 / 18.4 & \textcolor{red}{\textbf{65.9}} / \textcolor{red}{\textbf{67.5}} & \textcolor{blue}{\textbf{42.2}}
                & 10.8 / 11.5 & \textcolor{red}{\textbf{45.6}} / \textcolor{red}{\textbf{46.5}} & \textcolor{blue}{\textbf{28.6}}
                & 7.2  / 8.4  & \textcolor{red}{\textbf{32.0}} / \textcolor{red}{\textbf{36.2}} & \textcolor{blue}{\textbf{20.9}} \\
                & ~~+TDE~\cite{tang2020unbiased}$_{\textit{CVPR'20}}$
                & 26.2 / 29.6 & 44.8 / 49.2 & 37.5
                & 15.2 / 17.5 & 28.8 / 32.0 & 23.4
                & 9.5  / 11.4 & 17.3 / 20.9 & 14.8 \\
                & ~~+PCPL~\cite{yan2020pcpl}$_{\textit{MM'20}}$
                & 22.8 / 24.5 & \textcolor{blue}{\textbf{56.9}} / \textcolor{blue}{\textbf{58.7}} & 40.7
                & 15.2 / 16.1 & \textcolor{blue}{\textbf{40.6}} / \textcolor{blue}{\textbf{41.7}} & 28.4
                & 10.8 / 12.6 & 26.6 / 30.3 & 20.1 \\
                & ~~+CogTree~\cite{yu2021cogtree}$_{\textit{IJCAI'21}}$
                & 27.6 / 29.7 & 44.0 / 45.4 & 36.7
                & 18.8 / 19.9 & 30.9 / 31.7 & 25.3
                & 10.4 / 12.1 & 18.2 / 20.4 & 15.3 \\
                & ~~+DLFE~\cite{chiou2021recovering}$_{\textit{MM'21}}$
                & 25.3 / 27.1 & 51.8 / 53.5 & 39.4
                & 18.9 / 20.0 & 33.5 / 34.6 & 26.8
                & 11.8 / 13.8 & 22.7 / 26.3 & 18.7 \\
                & ~~+BPL-SA~\cite{guo2021general}$_{\textit{ICCV'21}}$
                & \textcolor{blue}{\textbf{30.6}} / \textcolor{blue}{\textbf{32.6}} & 50.0 / 51.8 & 41.3
                & \textcolor{red}{\textbf{20.1}} / \textcolor{blue}{\textbf{21.2}} & 34.0 / 35.0 & 27.6
                & \textcolor{red}{\textbf{13.5}} / \textcolor{red}{\textbf{15.7}} & 21.7 / 25.5 & 19.1 \\
                & \cellcolor{mygray-bg}{~~+\textbf{NICE (ours)}}
                & \cellcolor{mygray-bg}{\textcolor{red}{\textbf{30.7}}} / \cellcolor{mygray-bg}{\textcolor{red}{\textbf{33.0}}} &  \cellcolor{mygray-bg}{55.0} / \cellcolor{mygray-bg}{56.9} & \cellcolor{mygray-bg}{{\textcolor{red}{\textbf{43.9}}}}& \cellcolor{mygray-bg}\textcolor{blue}{\textbf{19.9}} / \cellcolor{mygray-bg}{\textcolor{red}{\textbf{21.3}}}& \cellcolor{mygray-bg}{37.8} / \cellcolor{mygray-bg}{39.0} &
                \cellcolor{mygray-bg}{\textcolor{red}{\textbf{29.5}}}	& \cellcolor{mygray-bg}\textcolor{blue}{\textbf{11.9}} / \cellcolor{mygray-bg}\textcolor{blue}{\textbf{14.1}}	&  \cellcolor{mygray-bg}\textcolor{blue}{\textbf{27.0}} / \cellcolor{mygray-bg}\textcolor{blue}{\textbf{30.8}} & \cellcolor{mygray-bg}{\textcolor{red}{\textbf{\textbf{21.0}}}} \\
                \hline
            \end{tabular}
        }
    \vspace{-0.5em}
    \caption{Performance (\%) of state-of-the art SGG models on three SGG tasks. ``B" denotes the backbone of object detector (Faster R-CNN~\cite{ren2015faster}) in each SGG model: \ie, VGG-16~\cite{simonyan2014very} and ResNeXt-101-FPN~\cite{lin2017feature}. ``Mean" is the average of mR@50/100 and R@50/100.  The \textcolor{red}{\textbf{best}} and \textcolor{blue}{\textbf{second best}} methods under each setting are marked according to formats.}
    \label{tab:compare_with_sota}
\end{table*}
\addtolength{\tabcolsep}{2pt}

\section{Experiments} \label{sec:4}
\subsection{Experimental Settings and Details}

\textbf{Dataset.} We conducted all experiments on the Challenging VG dataset~\cite{krishna2017visual}. It contains 108,073 images in total. In this paper, we followed widely-used splits~\cite{xu2017scene}, which keep the 150 most frequent object categories and the 50 most frequent predicate categories. Specifically, 70\% of images are training set and 30\% of images are test set. Following~\cite{zellers2018neural}, we sampled 5,000 images from the training set as the val set. Besides, we followed~\cite{liu2019large} to divide all predicate categories into three parts based on the number of samples in training set: head ($>$10k), body (0.5k$\sim$10k), and tail ($<$0.5k).

\textbf{Tasks.} We evaluated NICE on three SGG tasks~\cite{xu2017scene}: 1) \emph{Predicate Classification} (\textbf{PredCls}): Given the ground-truth objects with labels, we need to only predict pairwise predicate categories. 2) \emph{Scene Graph Classification} (\textbf{SGCls}): Given the ground-truth object bounding boxes, we need to predict both the object categories and predicate categories. 3) \emph{Scene Graph Generation} (\textbf{SGGen}): Given an image, we need to detect all object bounding boxes, and predict both the object categories and predicate categories.

\textbf{Metrics.} We evaluated all results on three metrics: 1) \emph{Recall@K} (\textbf{R@K}): It calculates the proportion of top-K confident predicted relation triplets that are in the ground-truth. Following prior works, we used $K = \{50, 100\}$. 2) \emph{mean Recall@K} (\textbf{mR@K}): It calculates the recall for each predicate category separately, and then averages R@K over all predicates, \ie, it puts relatively more emphasis on the tail categories. 3) \textbf{Mean}: It is the mean of all mR@K and R@K scores. R@K favors head predicates, while mR@K favors tail ones. Thus, it is a comprehensive metric that can better reflect model performance on different predicates.

\textbf{Implementation Details.} It is described in the appendix.
% \textbf{NICE Training Details.} In Neg-NSD, we used model Motifs~\cite{zellers2018neural} as OOD detection model $\mathtt{F}_{\text{sgg}}^n$. The training settings (\eg, learning rate and batch size) follow the same settings of~\cite{tang2020unbiased} under PredCls task, except that it was trained with only foreground samples. In Pos-NSD, we used a pretrained Motifs~\cite{zellers2018neural} provided by~\cite{tang2020unbiased} as $\mathtt{F}_{\text{sgg}}^p$ to extract triplet features (cf. $\bm{h}_i^k$ in Eq.~\eqref{eq:4}) under PredCls task. The number of divided subsets was set to 4. In NSC, the $a$, $b$ and $c$ were set to 1, 0, and 10, respectively. Note that although we used two Motifs models (one is an off-the-shelf model) in NICE, we only need to try NICE for one time, and then we can use the obtained cleaner annotations for any SGG models.

% \textbf{SGG Training Details.} Since NICE is a model-agnostic strategy, thus, for different baselines (\eg, Motifs~\cite{zellers2018neural} and VCTree~\cite{tang2019learning}), we followed their respective configurations\footnote{We utilized the SGG benchmark provided by~\cite{tang2020unbiased} for all baselines.}.

\subsection{Comparisons with State-of-the-Arts}

\textbf{Settings.} Since NICE is a model-agnostic strategy, it can be seamlessly incorporated into any advanced SGG model. In this section, we equipped NICE into two baselines: \textbf{Motifs}~\cite{zellers2018neural} and \textbf{VCTree}~\cite{tang2019learning}, and compared them with the state-of-the-art SGG methods. According to the generalization of these methods, we group them into two categories: 1) \textbf{TDE}~\cite{tang2020unbiased}, \textbf{PCPL}~\cite{yan2020pcpl}, \textbf{CogTree}~\cite{yu2021cogtree}, \textbf{DLFE}~\cite{chiou2021recovering}, and \textbf{BPL-SA}~\cite{guo2021general}. These methods are all model-agnostic SGG debiasing strategies. For fair comparisons, we also reported their performance on the Motifs and VCTree baselines. 2) \textbf{KERN}~\cite{chen2019knowledge}, \textbf{G-RCNN}~\cite{yang2018graph}, \textbf{MSDN}~\cite{li2017scene}, \textbf{BGNN}~\cite{li2021bipartite}, and \textbf{DT2-ACBS}~\cite{desai2021learning}. These methods are specifically designed SGG models. All results are reported in Table~\ref{tab:compare_with_sota}.

\textbf{Results.} From the results in Table~\ref{tab:compare_with_sota}, we have the following observations: 1) Compared to the two strong baselines (\ie, Motifs and VCTree), our NICE can consistently improve model performance on metric mR@K over all three tasks (\eg, 6.7\% $\sim$ 14.5\% and 4.7\% $\sim$ 14.6\% absolute gains on metric mR@100 over Motifs and VCTree, respectively). 2) Compared to other state-of-the-art model-agnostic debiasing strategies, NICE can not only always achieve top performance on mR@K metrics, but also keep relatively high performance on R@K metrics, \ie, NICE can improve the tail categories performance significantly, and maintain good performance on head categories. Thus, NICE can realize a better trade-off between accuracy among different predicate categories, and always achieve the best mean scores.

\subsection{Ablation Studies}

\subsubsection{Ablation Studies on Neg-NSD}\label{sec:4.3.1}

\textbf{Hyperparameter Settings of Neg-NSD.} The hyperparameter in Neg-NSD is the threshold $\theta$ for the confidence score (cf. Eq.~\eqref{eq:1}). Particularly, when the threshold $\theta$ for one category is set to $100\%$, which means we never assign this category as pseudo labels. Without loss of generality, we choose three representative hyperparameter settings, \ie, we mine missing annotated triplets on 1) all predicate categories, or 2) only body and tail categories, or 3) only tail categories. The thresholds $\theta$ for corresponding head, body, and tail categories were set as $95\%$, $90\%$, and $60\%$, respectively. To disentangle the influence of the other two modules (\ie, Pos-NSD and NSC), we directly use the outputs of Neg-NSD and pristine positive samples for SGG training.

\begin{figure*}[htbp]
    \begin{minipage}[c]{0.43\linewidth}
        \centering
        \captionsetup{type=table} %% tell latex to change to table
        %%%%%%%%%%%%%%%%%%%%% Ablation - 1 %%%%%%%%%%%%%%%%%%%%%%\
        % \vspace{-1em}
        \subfloat[The ablation studies on mining missing annotated triplets in head, body and tail categories in Neg-NSD, respectively.]{
        \tablestyle{5pt}{1.05}\begin{tabular}{|ccc|c|c|c|}
        \hline
        \multicolumn{3}{|c|}{Categories} & \multicolumn{3}{c|}{PredCls} \\
        \hline
        \footnotesize{head} & \footnotesize{body} & \footnotesize{tail} & \footnotesize{mR@50/100} & \footnotesize{R@50/100}  & \footnotesize{Mean} \\
        \hline
        \Checkmark & \Checkmark & \Checkmark & 22.2 / 25.8  & 58.5 / 62.5  & 42.3 \\
        \XSolidBrush & \Checkmark & \Checkmark & 23.7 / 26.2  & 59.5 / 62.3  & 42.9 \\
        % \Checkmark & \XSolidBrush & \XSolidBrush & 16.4 / 17.8  & 66.2 / 68.0  & 42.1 \\
        \XSolidBrush & \XSolidBrush & \Checkmark & 23.3 / 25.2  & 62.3 / 64.5  & 43.8 \\
        \hline
        \end{tabular}}
        \vspace{-1em}

        %%%%%%%%%%%%%%%%%%%%% Ablation - 2 %%%%%%%%%%%%%%%%%%%%%%
        \subfloat[The ablation studies on different cutoff distance $d_c$ in Pos-NSD for head, body and tail predicates respectively.]{
        \tablestyle{5pt}{1.05}\begin{tabular}{|ccc|c|c|c|}
        \hline
        \multicolumn{3}{|c|}{Size of $d_c$} & \multicolumn{3}{c|}{PredCls} \\
        \hline
        \footnotesize{head} & \footnotesize{body} & \footnotesize{tail}  & \footnotesize{mR@50/100}   & \footnotesize{R@50/100}  & \footnotesize{Mean} \\
        \hline
        $\mathcal{L}$ & $\mathcal{L}$ & $\mathcal{L}$ & 19.9 / 21.5  & 64.0 / 65.7  & 42.8 \\
        % \hline
        $\mathcal{M}$ & $\mathcal{M}$ & $\mathcal{M}$ & 21.0 / 22.8  & 62.2 / 64.0  & 42.5 \\
        $\mathcal{S}$ & $\mathcal{S}$ & $\mathcal{S}$ & 21.6 / 23.4  & 61.1 / 62.8  & 42.2 \\
        % \hline
        $\mathcal{S}$ & $\mathcal{M}$ & $\mathcal{L}$ & 23.3 / 25.2  & 59.6 / 61.3  & 42.4 \\
        $\mathcal{L}$ & $\mathcal{M}$ & $\mathcal{S}$ & 18.6 / 20.1  & 64.4 / 66.1  & 42.3 \\
        \hline
         \end{tabular}}
        \vspace{-1em}

        %%%%%%%%%%%%%%%%%%%%% Ablation - 3 %%%%%%%%%%%%%%%%%%%%%%
        \subfloat[The ablation studies on different $K$ in wKNN.]{
		\tablestyle{6pt}{1.05}\begin{tabular}{|c|c|c|c|}
			\hline
            \multirow{2}{*}{$K$} & \multicolumn{3}{c|}{PredCls} \\
            \cline{2-4}
            & \footnotesize{mR@50/100} & \footnotesize{R@50/100}  & \footnotesize{Mean} \\
            \hline
            $K$ = 1 & 23.3 / 25.0 & 59.3 / 60.9 & 42.1 \\
            $K$ = 3 & 23.3 / 25.2 & 59.6 / 61.3 & 42.4 \\
            $K$ = 5 & 22.9 / 24.7 & 59.8 / 61.5 & 42.2 \\
            \hline
	    \end{tabular}}
	    \vspace{-1em}
    	\caption{Ablation studies on the influence of different hyperparameters of each component of NICE. Motifs~\cite{zellers2018neural} is the baseline model which are used in all experiments.}
    	\label{tab:ablations}
    \end{minipage} \hfill
    \begin{minipage}[c]{0.55\linewidth}
        %%%%%%%%%%%%%%% t-SNE visualization figure %%%%%%%%%%%%%
        \centering
        \includegraphics[width=\linewidth]{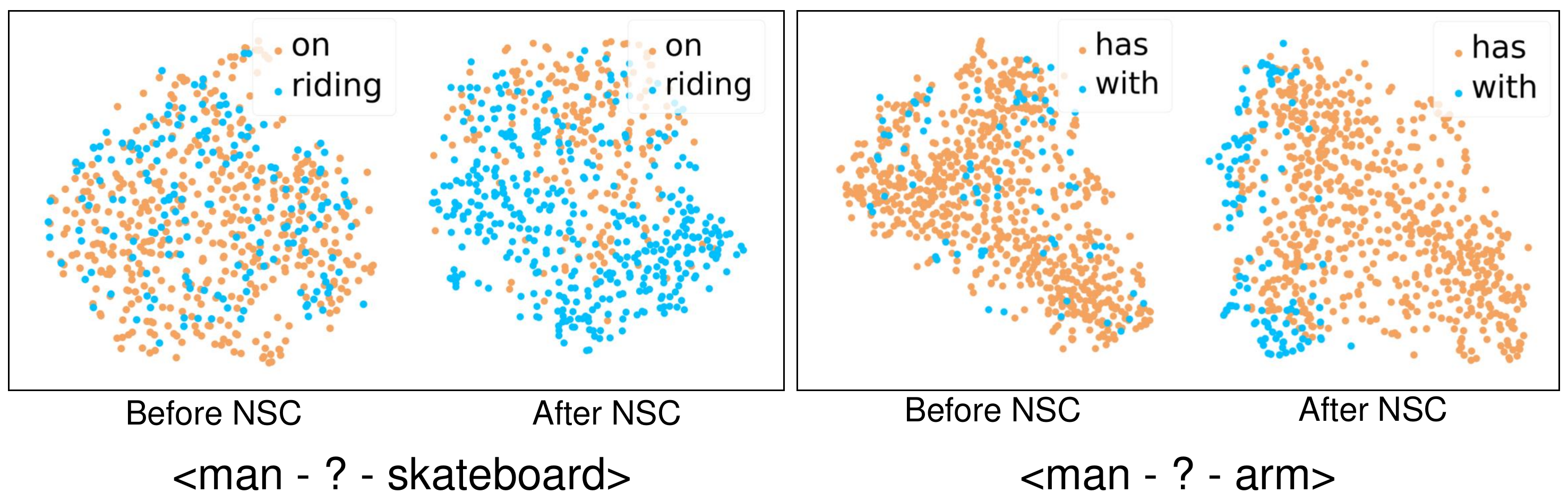}
        \vspace{-2em}
        \caption{The t-SNE visualization of randomly sampled instances of different triplet categories on the feature space before and after NSC.}
        \label{fig:t-sne}

        %%%%%%%%%%%%%%%%%%% visualization figure %%%%%%%%%%%%%
        \includegraphics[width=\linewidth]{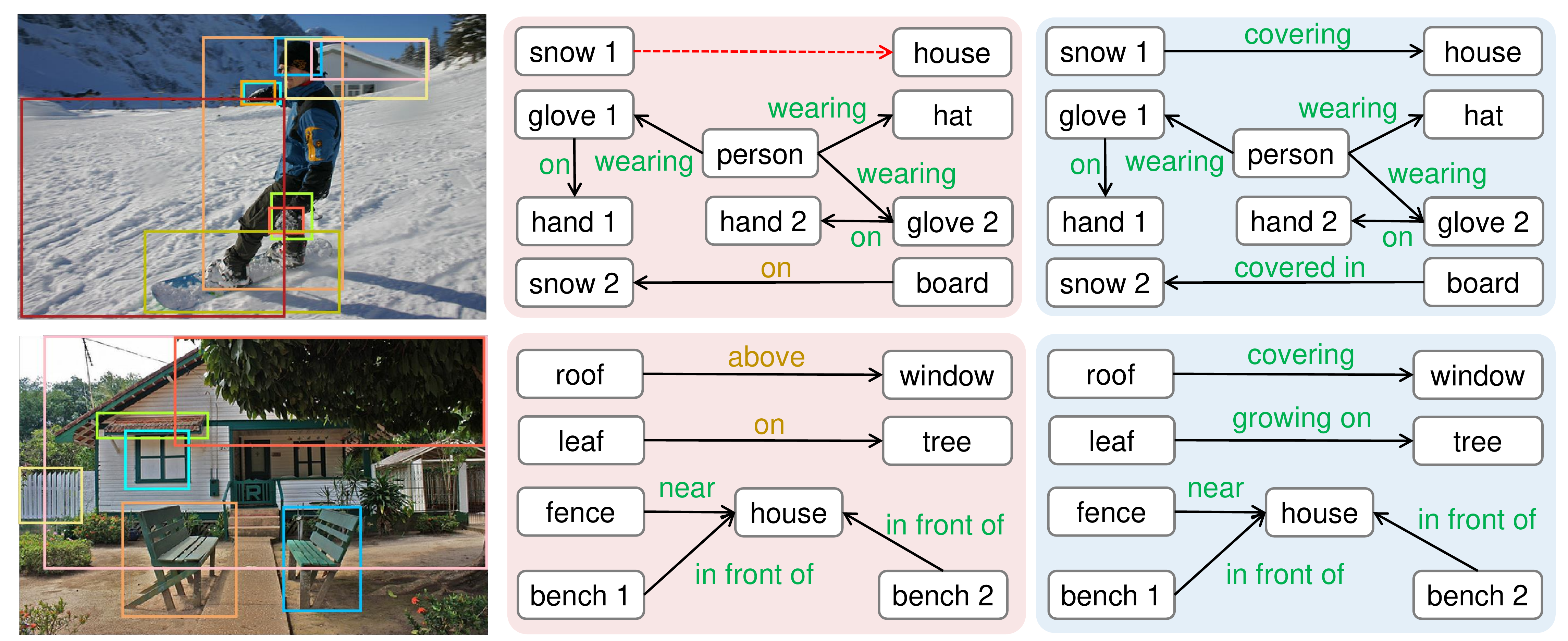}
        \vspace{-2em}
        \caption{Scene graphs generated by Motifs (left) and Motifs+NICE (right) on PredCls. \textcolor{mygreen}{\textbf{Green}} predicates are correct (\ie, GT), and \textcolor{brown}{\textbf{brown}} predicates are more coarse-grained (not in GT). \textcolor{red}{\textbf{Red}} dashed arrows are the relations that are not detected. Only detected boxes overlapped with GT are shown.}
        \label{fig:vis}
    \end{minipage}
\end{figure*}

\emph{Results.} From the results in Table~\ref{tab:ablations}(a), we can observe: 1) Different threshold settings have a slight influence on the mR@K metrics, but relatively more influence on the R@K metrics. 2) When only mining the missing tail predicates in Neg-NSD, the model gains the best performance.

\noindent\textbf{Effectiveness of Neg-NSD.} We evaluated the effectiveness of the Neg-NSD by using the same refined samples by Neg-NSD and pristine positive samples for SGG training.

\emph{Results.} All results are reported in Table~\ref{tab:component}. Compared to the baseline model (\#~1), the Neg-PSD module (\#~2) can significantly improve model performance on mR@K metrics (\eg, $25.2\%$ vs. $17.8\%$ in mR@100), which proves that these harvested ``positive" samples (noisy negative samples with pseudo labels) are indeed beneficial for SGG training.

\begin{table}[!t]
\addtolength{\tabcolsep}{-2.5pt}
\renewcommand{\arraystretch}{1.05}
    \centering
    \scalebox{0.95}{
        \begin{tabular}{|l|ccc|c|c|c|c|}
        \hline
        & \multicolumn{3}{c|}{Components} & \multicolumn{3}{c|}{PredCls} \\
        \cline{2-7}
        \# & \small{N-NSD} & \small{P-NSD} & \small{NSC}     & \small{mR@50/100} & \small{R@50/100}  & \small{Mean} \\
        \hline
        1 & \XSolidBrush & \XSolidBrush & \XSolidBrush  & 16.5 / 17.8    & 65.5 / 67.2  & 41.8 \\
        2 & \Checkmark & \XSolidBrush & \XSolidBrush  & 23.3 /	25.2  &  62.3 / 64.5  & 43.8 \\
        3 & \XSolidBrush & \Checkmark & \XSolidBrush & 20.3 / 22.0    & 57.6 / 59.2  & 39.8 \\
        4 & \Checkmark & \Checkmark & \XSolidBrush & 20.4 / 23.4 & 56.7 / 61.4  & 40.5 \\
        5 & \XSolidBrush & \Checkmark & \Checkmark & 23.3 / 25.2  & 59.6 / 61.3  & 42.4 \\
        6 & \Checkmark & \Checkmark & \Checkmark & 29.9 / 32.3  & 55.1 / 57.2  & 43.6 \\
        \hline
        \end{tabular}%
    }
    \vspace{-0.5em}
    \caption{Ablation studies on each component of NICE. ``\#" is the line number. The baseline model (\#~1) is Motifs~\cite{zellers2018neural}.}
    \label{tab:component}%
\addtolength{\tabcolsep}{2.5pt}
\end{table}

\subsubsection{Ablation Studies on Pos-NSD}

\noindent\textbf{Hyperparameter Settings of Pos-NSD.} The hyperparameter in Pos-NSD is the cutoff distance $d_c$ ranked at $\alpha$\% for different categories (cf. Eq.~\eqref{eq:5}). As mentioned in Sec.~\ref{sec:3.2}, different $d_c$ directly influence the clustering results of each predicate category, and smaller $d_c$ is more suitable for predicates with multiple semantic meanings. Thus, without loss of generality, we choose five typical settings for different categories, including ($\mathcal{L}, \mathcal{L}, \mathcal{L}$),
($\mathcal{M}, \mathcal{M}, \mathcal{M}$),  and so on. The $\mathcal{L}$, $\mathcal{M}$, $\mathcal{S}$ denote the $d_c$ are large, medium, and small. In our experiments, their corresponding $\alpha\%$ were set to $50.0\%$, $25.0\%$ and $12.5\%$. Similarly, we disentangle the influence of Neg-NSD, and use pristine negative samples and refined positive samples (outputs of NSC) for SGG training.

\emph{Results.} When the cutoff distance $d_c$ for all predicate categories are set to large or small or from large to small, the model achieves relatively worse results. These results are also consistent with our expectation, \ie, for the predicate categories with multiple semantic meanings (head categories), small $d_c$ is better for noisy sample detection. Instead, for predicate categories with unique semantic meaning (tail categories), larger $d_c$ is better. Thus, we utilize the ($\mathcal{S}, \mathcal{M}, \mathcal{L}$) setting for all experiments.

\noindent\textbf{Effectiveness of Pos-NSD.} We evaluated the effectiveness of Pos-NSD by just using clean positive samples detected from $\bm{\mathcal{T}}^+$ with pristine negative samples or from the new positive samples set ${\widetilde {\bm{\mathcal{T}}}^+}$ for SGG training.

\emph{Results.} As shown in Table~\ref{tab:component} (\# 3), the single Pos-NSD component can still improve mR@K metrics by using much fewer positive training samples. Besides, baseline can also be exceeded on mR@K (\# 4) after Neg-NSD and Pos-NSD alone with fewer training samples. It proves that the clean subset divided by Pos-NSD is better for SGG training, and numerous noisy positive samples actually hurt performance.

\subsubsection{Ablation Studies on NSC}

\noindent\textbf{Hyperparameter of NSC.} The hyperparameter of NSC is the $K$ in wKNN. We investigated $K=\{1,3,5\}$. All results are reported in Table~\ref{tab:ablations}(c). From the results, we can observe that the SGG performance is robust to different $K$. To better trade-off different metrics, we set $K$ to 3.

\noindent\textbf{Effectiveness of NSC.} Based on Pos-NSD, NSC replaces the original labels of the noisy positive samples with cleaner labels. As shown in Table~\ref{tab:component}, NSC markedly improves SGG performance on both mR@K and R@K (\#~5 vs. \#~3), \ie, NSC always reassigns better labels to noisy samples.

\noindent\textbf{T-SNE Visualizations.} We visualized the t-SNE distributions of features of $\langle$\texttt{man}-\texttt{on/riding}-\texttt{skateboard}$\rangle$ as well as $\langle$\texttt{man}-\texttt{has/with}-\texttt{arm}$\rangle$ on the feature space before and after the NSC in Figure~\ref{fig:t-sne}. And the features are from the previous layer of
classification, like P-NSD. The former is a pair of predicates of different granularity (common-prone), and the latter is a pair of predicates of the same granularity (synonym-random). As shown in Figure~\ref{fig:t-sne}, NSC can help to alleviate the inconsistency of ground-truths, \ie, similar visual patterns always have more consistent ground-truth predicate annotations, which is beneficial for SGG training.
\subsubsection{Qualitative Results}

Figure~\ref{fig:vis} demonstrates some qualitative results generated by Motifs and Motifs+NICE. From Figure~\ref{fig:vis}, we can observe that NICE can not only help to detect more missing false positive predicates (\eg, \texttt{covering}), but also more fine-grained and informative predicates (\eg, \texttt{growing on} vs. \texttt{on}, and \texttt{covering} vs. \texttt{above}).

\section{Conclusions and Limitations} \label{conclusion}

In this paper, we argued that two plausible assumptions about the ground-truth annotations are inapplicable for existing SGG datasets. To this end, we reformulated SGG as a noisy label learning problem and proposed a novel model-agnostic noisy label correction strategy: NICE. NICE can not only detect noisy samples, but also reassign better predicate labels to them. We have validated the effectiveness of each part of NICE through extensive experiments.

\textbf{Limitations.}
Although NICE can mine some potential unseen triplets in Neg-NSD, there is no guarantee that the harvested triplets must be reasonable. Meanwhile, some hyperparameters have different impacts on each predicate category, which makes it difficult to achieve the best trade-off between different metrics (cf. mR@K \& R@K in Table~\ref{tab:ablations}).

\footnotesize \noindent\textbf{Acknowledgement} This work was supported by National Key Research \& Development Project of China (2021ZD0110700), National Natural Science Foundation of China (U19B2043, 61976185), Zhejiang Natural Science Foundation (LR19F020002), Zhejiang Innovation Foundation (2019R52002), and Fundamental Research Funds for Central Universities.

% Moving forward, we plan to : 1) design a more effective SGG model to benefit more from the NICE refined samples; 2) extend NICE to other tasks (\eg, video SGG), which also suffers from noisy annotations.


{\small
\begin{thebibliography}{10}\itemsep=-1pt

\bibitem{carion2020end}
Nicolas Carion, Francisco Massa, Gabriel Synnaeve, Nicolas Usunier, Alexander
  Kirillov, and Sergey Zagoruyko.
\newblock End-to-end object detection with transformers.
\newblock In {\em ECCV}, pages 213--229, 2020.

\bibitem{chen2019counterfactual}
Long Chen, Hanwang Zhang, Jun Xiao, Xiangnan He, Shiliang Pu, and Shih-Fu
  Chang.
\newblock Counterfactual critic multi-agent training for scene graph
  generation.
\newblock In {\em ICCV}, pages 4613--4623, 2019.

\bibitem{chen2019knowledge}
Tianshui Chen, Weihao Yu, Riquan Chen, and Liang Lin.
\newblock Knowledge-embedded routing network for scene graph generation.
\newblock In {\em CVPR}, pages 6163--6171, 2019.

\bibitem{chiou2021recovering}
Meng-Jiun Chiou, Henghui Ding, Hanshu Yan, Changhu Wang, Roger Zimmermann, and
  Jiashi Feng.
\newblock Recovering the unbiased scene graphs from the biased ones.
\newblock In {\em ACM MM}, 2021.

\bibitem{desai2021learning}
Alakh Desai, Tz-Ying Wu, Subarna Tripathi, and Nuno Vasconcelos.
\newblock Learning of visual relations: The devil is in the tails.
\newblock In {\em ICCV}, pages 15404--15413, 2021.

\bibitem{devries2018learning}
Terrance DeVries and Graham~W Taylor.
\newblock Learning confidence for out-of-distribution detection in neural
  networks.
\newblock In {\em arXiv}, 2018.

\bibitem{gao2022classification}
Kaifeng Gao, Long Chen, Yulei Niu, Jian Shao, and Xiao Jun.
\newblock Classification-then-grounding: Reformulating video scene graphs as
  temporal bipartite graphs.
\newblock In {\em CVPR}, 2022.

\bibitem{goldberger2016training}
Jacob Goldberger and Ehud Ben-Reuven.
\newblock Training deep neural-networks using a noise adaptation layer.
\newblock In {\em ICLR}, 2017.

\bibitem{guo2018curriculumnet}
Sheng Guo, Weilin Huang, Haozhi Zhang, Chenfan Zhuang, Dengke Dong, Matthew~R
  Scott, and Dinglong Huang.
\newblock Curriculumnet: Weakly supervised learning from large-scale web
  images.
\newblock In {\em ECCV}, pages 135--150, 2018.

\bibitem{guo2021general}
Yuyu Guo, Lianli Gao, Xuanhan Wang, Yuxuan Hu, Xing Xu, Xu Lu, Heng~Tao Shen,
  and Jingkuan Song.
\newblock From general to specific: Informative scene graph generation via
  balance adjustment.
\newblock In {\em ICCV}, pages 16383--16392, 2021.

\bibitem{hechenbichler2004weighted}
Klaus Hechenbichler and Klaus Schliep.
\newblock Weighted k-nearest-neighbor techniques and ordinal classification.
\newblock 2004.

\bibitem{hendrycks2016baseline}
Dan Hendrycks and Kevin Gimpel.
\newblock A baseline for detecting misclassified and out-of-distribution
  examples in neural networks.
\newblock {\em arXiv}, 2016.

\bibitem{jiang2018mentornet}
Lu Jiang, Zhengyuan Zhou, Thomas Leung, Li-Jia Li, and Li Fei-Fei.
\newblock Mentornet: Learning data-driven curriculum for very deep neural
  networks on corrupted labels.
\newblock In {\em ICML}, pages 2304--2313, 2018.

\bibitem{knyazev2020graph}
Boris Knyazev, Harm de Vries, C{\u{a}}t{\u{a}}lina Cangea, Graham~W Taylor,
  Aaron Courville, and Eugene Belilovsky.
\newblock Graph density-aware losses for novel compositions in scene graph
  generation.
\newblock In {\em arXiv}, 2020.

\bibitem{krishna2017visual}
Ranjay Krishna, Yuke Zhu, Oliver Groth, Justin Johnson, Kenji Hata, Joshua
  Kravitz, Stephanie Chen, Yannis Kalantidis, Li-Jia Li, David~A Shamma, et~al.
\newblock Visual genome: Connecting language and vision using crowdsourced
  dense image annotations.
\newblock {\em IJCV}, 2017.

\bibitem{lee2018cleannet}
Kuang-Huei Lee, Xiaodong He, Lei Zhang, and Linjun Yang.
\newblock Cleannet: Transfer learning for scalable image classifier training
  with label noise.
\newblock In {\em CVPR}, pages 5447--5456, 2018.

\bibitem{li2021bipartite}
Rongjie Li, Songyang Zhang, Bo Wan, and Xuming He.
\newblock Bipartite graph network with adaptive message passing for unbiased
  scene graph generation.
\newblock In {\em CVPR}, pages 11109--11119, 2021.

\bibitem{li2017vip}
Yikang Li, Wanli Ouyang, Xiaogang Wang, and Xiao'ou Tang.
\newblock Vip-cnn: Visual phrase guided convolutional neural network.
\newblock In {\em CVPR}, pages 1347--1356, 2017.

\bibitem{li2017scene}
Yikang Li, Wanli Ouyang, Bolei Zhou, Kun Wang, and Xiaogang Wang.
\newblock Scene graph generation from objects, phrases and region captions.
\newblock In {\em ICCV}, pages 1261--1270, 2017.

\bibitem{li2017learning}
Yuncheng Li, Jianchao Yang, Yale Song, Liangliang Cao, Jiebo Luo, and Li-Jia
  Li.
\newblock Learning from noisy labels with distillation.
\newblock In {\em ICCV}, pages 1910--1918, 2017.

\bibitem{lin2017feature}
Tsung-Yi Lin, Piotr Doll{\'a}r, Ross Girshick, Kaiming He, Bharath Hariharan,
  and Serge Belongie.
\newblock Feature pyramid networks for object detection.
\newblock In {\em CVPR}, pages 2117--2125, 2017.

\bibitem{lin2017focal}
Tsung-Yi Lin, Priya Goyal, Ross Girshick, Kaiming He, and Piotr Doll{\'a}r.
\newblock Focal loss for dense object detection.
\newblock In {\em ICCV}, pages 2980--2988, 2017.

\bibitem{lin2020gps}
Xin Lin, Changxing Ding, Jinquan Zeng, and Dacheng Tao.
\newblock Gps-net: Graph property sensing network for scene graph generation.
\newblock In {\em CVPR}, pages 3746--3753, 2020.

\bibitem{liu2019large}
Ziwei Liu, Zhongqi Miao, Xiaohang Zhan, Jiayun Wang, Boqing Gong, and Stella~X
  Yu.
\newblock Large-scale long-tailed recognition in an open world.
\newblock In {\em CVPR}, pages 2537--2546, 2019.

\bibitem{lu2016visual}
Cewu Lu, Ranjay Krishna, Michael Bernstein, and Li Fei-Fei.
\newblock Visual relationship detection with language priors.
\newblock In {\em ECCV}, pages 852--869, 2016.

\bibitem{ma2018dimensionality}
Xingjun Ma, Yisen Wang, Michael~E Houle, Shuo Zhou, Sarah Erfani, Shutao Xia,
  Sudanthi Wijewickrema, and James Bailey.
\newblock Dimensionality-driven learning with noisy labels.
\newblock In {\em ICML}, pages 3355--3364, 2018.

\bibitem{ren2018learning}
Mengye Ren, Wenyuan Zeng, Bin Yang, and Raquel Urtasun.
\newblock Learning to reweight examples for robust deep learning.
\newblock In {\em ICML}, pages 4334--4343, 2018.

\bibitem{ren2015faster}
Shaoqing Ren, Kaiming He, Ross Girshick, and Jian Sun.
\newblock Faster r-cnn: Towards real-time object detection with region proposal
  networks.
\newblock In {\em NeurIPS}, pages 91--99, 2015.

\bibitem{rodriguez2014clustering}
Alex Rodriguez and Alessandro Laio.
\newblock Clustering by fast search and find of density peaks.
\newblock {\em Science}, pages 1492--1496, 2014.

\bibitem{simonyan2014very}
Karen Simonyan and Andrew Zisserman.
\newblock Very deep convolutional networks for large-scale image recognition.
\newblock {\em arXiv}, 2014.

\bibitem{tang2020unbiased}
Kaihua Tang, Yulei Niu, Jianqiang Huang, Jiaxin Shi, and Hanwang Zhang.
\newblock Unbiased scene graph generation from biased training.
\newblock In {\em CVPR}, pages 3716--3725, 2020.

\bibitem{tang2019learning}
Kaihua Tang, Hanwang Zhang, Baoyuan Wu, Wenhan Luo, and Wei Liu.
\newblock Learning to compose dynamic tree structures for visual contexts.
\newblock In {\em CVPR}, pages 6619--6628, 2019.

\bibitem{vahdat2017toward}
Arash Vahdat.
\newblock Toward robustness against label noise in training deep discriminative
  neural networks.
\newblock In {\em NeurIPS}, 2017.

\bibitem{wang2020tackling}
Tzu-Jui~Julius Wang, Selen Pehlivan, and Jorma Laaksonen.
\newblock Tackling the unannotated: Scene graph generation with bias-reduced
  models.
\newblock In {\em BMVC}, 2020.

\bibitem{wang2022crossformer}
Wenxiao Wang, Lu Yao, Long Chen, Binbin Lin, Deng Cai, Xiaofei He, and Wei Liu.
\newblock Crossformer: A versatile vision transformer hinging on cross-scale
  attention.
\newblock In {\em ICLR}, 2022.

\bibitem{wang2019symmetric}
Yisen Wang, Xingjun Ma, Zaiyi Chen, Yuan Luo, Jinfeng Yi, and James Bailey.
\newblock Symmetric cross entropy for robust learning with noisy labels.
\newblock In {\em ICCV}, pages 322--330, 2019.

\bibitem{xu2017scene}
Danfei Xu, Yuke Zhu, Christopher~B Choy, and Li Fei-Fei.
\newblock Scene graph generation by iterative message passing.
\newblock In {\em CVPR}, pages 5410--5419, 2017.

\bibitem{xu2019l_dmi}
Yilun Xu, Peng Cao, Yuqing Kong, and Yizhou Wang.
\newblock L\_dmi: An information-theoretic noise-robust loss function.
\newblock In {\em arXiv}, 2019.

\bibitem{yan2020pcpl}
Shaotian Yan, Chen Shen, Zhongming Jin, Jianqiang Huang, Rongxin Jiang, Yaowu
  Chen, and Xian-Sheng Hua.
\newblock Pcpl: Predicate-correlation perception learning for unbiased scene
  graph generation.
\newblock In {\em ACM MM}, pages 265--273, 2020.

\bibitem{yang2018graph}
Jianwei Yang, Jiasen Lu, Stefan Lee, Dhruv Batra, and Devi Parikh.
\newblock Graph r-cnn for scene graph generation.
\newblock In {\em ECCV}, pages 670--685, 2018.

\bibitem{yu2021cogtree}
Jing Yu, Yuan Chai, Yue Hu, and Qi Wu.
\newblock Cogtree: Cognition tree loss for unbiased scene graph generation.
\newblock In {\em IJCAI}, 2021.

\bibitem{zellers2018neural}
Rowan Zellers, Mark Yatskar, Sam Thomson, and Yejin Choi.
\newblock Neural motifs: Scene graph parsing with global context.
\newblock In {\em CVPR}, pages 5831--5840, 2018.

\bibitem{zhang2017visual}
Hanwang Zhang, Zawlin Kyaw, Shih-Fu Chang, and Tat-Seng Chua.
\newblock Visual translation embedding network for visual relation detection.
\newblock In {\em CVPR}, pages 5532--5540, 2017.

\bibitem{zhang2018generalized}
Zhilu Zhang and Mert~R Sabuncu.
\newblock Generalized cross entropy loss for training deep neural networks with
  noisy labels.
\newblock In {\em NeurIPS}, 2018.

\end{thebibliography}
}
\end{document}